%% file: main.tex
\documentclass[10pt,conference]{IEEEtran}
\IEEEoverridecommandlockouts
\usepackage{cite}
\usepackage{amsmath,amssymb,amsfonts}
\usepackage{algorithmic}
\usepackage{graphicx}
\usepackage{textcomp}
\usepackage{balance}
\usepackage{url}

\usepackage{xcolor}

\def\BibTeX{{\rm B\kern-.05em{\sc i\kern-.025em b}\kern-.08em
    T\kern-.1667em\lower.7ex\hbox{E}\kern-.125emX}}

\begin{document}

\title{mdx: A Cloud Platform for Supporting Data Science and Cross-Disciplinary Research Collaborations}

\input{authors.tex}

\maketitle
\begin{abstract}
The growing amount of data and advances in data science have created a need for a new kind of cloud platform that provides users with flexibility, strong security, and the ability to couple with supercomputers and edge devices through high-performance networks. We have built such a nation-wide cloud platform, called ``mdx'' to meet this need. The mdx platform's virtualization service, jointly operated by 9 national universities and 2 national research institutes in Japan, launched in 2021, and more features are in development. Currently mdx is used by researchers in a wide variety of domains, including materials informatics, geo-spatial information science, life science, astronomical science, economics, social science, and computer science.  This paper provides an the overview of the mdx platform, details the motivation for its development, reports its current status, and outlines its future plans.
\end{abstract}


\section{Introduction}
\label{introduction}

Increases in the volume of data available, along with advances in tools for its analysis, are revolutionizing science, business, and society. Termed "the fourth paradigm shift"~\cite{hey2009fourth}, this revolution makes it possible for scientific discoveries to be driven by data in a bottom-up manner, as opposed to the top-down approach associated with traditional science, where discovery often comes from developing and deploying detailed theoretical models such as Newton's laws of motion. The data-driven approach of the fourth paradigm has become indispensable to both business and areas of scientific research such as materials science, life science, and weather prediction.  

Despite the power of data-driven approaches, there are still many fields where they are underutilized. A shortage of personnel with the requisite data science or computer science skills is one reason for this. Legacy systems that hinder systematic data extraction are another. Institutional and cultural barriers to data sharing among researchers are also a contributing factor in many situations, although there is a growing movement toward open science and open data~\cite{VICENTESAEZ2018428}. Thus, large gaps still exist in many research areas between their current state and what they could achieve by incorporating modern data-driven analysis. 

Research problems of fundamental importance and high value to society, such as controlling carbon emissions, creating smart cities, expediting drug discovery, and improving economic models are so complex that meaningful progress on them is only possible through interdisciplinary collaboration~\cite{datascience1}. 
Thus, facilitating cross-domain collaborations that can benefit from data-driven analysis has the potential to effect enormous positive change.  

 However, the diversity of needs that researchers have with regard to computing and storage resources makes pre-existing cloud platforms ill-suited for the role facilitating collaboration described above. Some collaborations handle small volumes data; others handle large amounts. Some needs big simulations and access to supercomputers; others need only laptop-level computing power but might rely on a shared environment such as Jupyter notebooks. This broad range of requirements led us to design and build a new type of cloud platform. 

This new platform, called "mdx", is now currently operated by 9 national universities and 2 national research institutes in Japan. In this paper we give an overview of the mdx platform. Significant  aspects of its development include the following: 

\begin{itemize}
      \item An architectural design pattern that allows users to set up their own virtualization computing platforms, thus building a nation-wide cloud platform for data science in a scalable manner.       
      \item Holistically integrated with supercomputers, edge devices through the high-performance network. 
      \item From the fall of 2021 supporting more than 50 projects that use mdx in various domains, including materials science, smart city-related projects, carbon neutrality, and medical data analysis. 
\end{itemize}

In this paper, we firstly describe motivations and list of requirement for the new platform in Section \ref{motivation}, and then in Section \ref{relatedwork} we describe relate work, including activities in Europe and the US. Section \ref{mdx} describes the overview of the mdx platform, and Section~\ref{status} describes the status of the project. We conclude with a summary and discussion of future directions for the work in Section~\ref{summary}. 

\section{Motivation}\label{motivation}

As described in the previous section, better large-scale coordination is essential to making progress in a number of important fundamental and applied research areas.  To meet this challenge, we have built a nation-wide cloud platform called ``mdx''. In this section, we list the requirements for this platform. We also describe why developing a new platform, rather than using supercomputers or existing commercial cloud services, was necessary. 

\subsection{Flexibility with Regard to Computing Environment }  Conventional supercomputers typically restrict users to using a pre-installed operating system (e.g. Windows, Ubuntu, CentOS, etc), certain software libraries (e.g. Python, Pytorch, Tensorflow, scikit-learn), and pre-installed device drivers (e.g. NVIDIA GPU driver). Such inflexibility presents a challenge to data scientists and domain experts, whose required computing environment varys widely depending on application and user preference. One data scientist might need specific versions of Pytorch, Tensorflow and NVIDIA GPU drivers for their deep learning applications. Another might need a Windows desktop server since they have a visualization software package that only runs on that operating system. To satisfy the multitudinous requirements in as wide a variety of application domains as possible, the mdx platform allows users to build their computing environments and install their required software packages by themselves.  

\subsection{Flexibility with Regard to Hosting Service}
Conventional supercomputers are not designed to flexibly accommodate users' hosting preferences because only log-in nodes and specific nodes are accessible from an external network. However, in various domains researchers need a web portal service that shares their data with the public or certain groups with privilege access.  For example, materials informatics researchers may need to share simulation results computed by the first-principle computation; a geo-spatial science or smart city project may need to share real-time streaming data from IoT (Internet-of-Things) sensors that are deployed in a city; and genome science researchers may need to share sensitive genome data only with permitted collaborators. User control over hosting is therefore an important feature of the mdx platform. 


\subsection{Agility Creating a Secure Virtual Environment} 
With respect to security, virtualization, computing resources, and storage, the ideal cloud platform is agile, allowing an individual or set of collaborating researchers to set up a secure and virtual environments. One way to accomplish this is by employing technologies such as L2VPN (Layer2 Virtual Private Network) to allow network-level virtualization. Computing resources and storage resources should be reconfigurable based on their prompt needs because data science applications typically require more GPU devices once they build up a new deep learning model. The cycle of software development and testing is iterative in most cases, so users' need for GPU devices is dynamic. Thus, giving users the ability to claim computing and storage devices on demand was an important when developing mdx.

\subsection{Seamless Coupling with Data Sources}  
Thanks to the advance of network technologies, various types of edges -- from small IoT sensors to large digital experimental equipment such as large-scale electroscopes -- are now connected and accessible through the Internet. 
To leverage data from such sensors, the ideal platform should provide functionality that allows users to seamlessly obtain such data without system/network configurations or programming effort. Ideally, users would be able use those data in a Kaggle manner, where users select their required data sources and can directly access the data without being able to copy the data to their computing environment.  

\subsection{Real-time Response} Real-time response is needed for applications such as real-time traffic accident identification, real-time anomaly detection, real-time disaster management such as identifying floods from satellite images, etc.  Current supercomputers are designed for batch-based job scheduling that does not guarantee job scheduling, so that they are not suitable for applications that need real-time response.  
In a BoF session at Supercomputing 2018~\cite{bofsc18realtimehpc}, researchers discussed how real-time supercomputers can be achieved.

\subsection{Fair Resource Provisioning and Pricing}
In order to offer computational resources and services to a wide range of users who varying greatly in how intensive their usage will be, the platform needs to provision these resources in a fair way. 
Moreover, for its sustainable service operations, the collecting of maintenance costs from users is necessary even though the platform is a non-profit service.

Fair resource provisioning and pricing in a non-profit way are key challenges, and there is an opportunity to learn from the experience of supercomputer facilities, which determined that when resources were allocated in a FIFO (first in, first out) manner, resources are typically occupied by limited users who have more funding.
To give more equal opportunities to a wide range of researchers regardless of whether they have big funding or not, the platform needs to introduce different types of resource scheduling and pricing mechanisms. 
Moreover, the pricing policy should be designed in a different manner than commercial cloud services so that the platform would accelerate interdisciplinary research collaborations. 

\subsection{Representative Motivating Applications}
\noindent 
\textbf{Materials Informatics:}
Materials informatics is an interdisciplinary research field that aims to solve materials science problems by using computing technologies such as computational simulation and machine learning. The materials informatics workflow requires (i) high-performance computing power and (ii) large-scale storage while ensuring (iii) strong secure network isolation. These requirements are not achieved on the whole solely by using a supercomputer or cloud.
First, in order to predict material features theoretically, extremely high-performance computing power is crucial. E.g., first-principles calculations may require the full scale of world-class supercomputers when solving large-scale molecular systems.  
Second, the empirical data from experimental instruments, e.g. electronic microscopes and synchrotron radiation instruments~\cite{spring-8}, often becomes very large due to the high-resolution images. Large storage and high-performance network from the instruments to the storage are necessary as well as computing power.
Third, strongly secure network isolation is crucial not only because the generated data are highly valuable but also because experimental instruments still use legacy software. Unlike commodity servers, such instruments often only support some specific old OS versions and are incompatible with the others.

\smallskip
\noindent
\textbf{Spatio-Temporal Analysis:}
Spatio-temporal analysis is a research area related to time-series data with geo-spatial dimension, such as traffic prediction and management. 
The widespread use of sensor/mobile devices and the recent advancement of machine learning technologies have given us new potential to analyze spatio-temporal data at scale as well as in real time. 
Such large-scale real-time spatio-temporal analysis requires (i) responsive high-performance computing nodes and (ii) a low-latency and secure network for these devices.
For example, in traffic prediction, which forecasts the future traffic condition based on previous conditions, inferences (e.g., by neural networks or behavioral simulations) need to be completed faster than real time, which requires very low-latency networks and responsive high-performance computing nodes. 
At the same time, the data must be handled through a secure network as they often include private information such as positional data for each vehicle or pedestrian.

\section{Related Work}\label{relatedwork}
This section summarizes projects and services relevant to our work.  

\smallskip
\noindent
\textbf{Platforms for Data Science:}
Several large-scale and federated computing platforms have been established recently to enhance the productivity of the data-science research.

The eXtreme Science and Engineering Discovery Environment (XSEDE)~\cite{xsede} provides scholars, researchers, and engineers for scientific researches with high-end computing resources and digital services, such as supercomputers, virtual machines, storage, and analytic/visualization tools, as well as collaborative support services for research and education. The XSEDE resources consist of multiple campus-based resources,  including Stampede2/Bridges2 supercomputers~\cite{stanzione2017stampede,brown2021bridges} and the Jetstream2 cloud system ~\cite{jetstream2}.

The European Open Science Cloud (EOSC)~\cite{eosc} provides a federated and multi-disciplinary research environment for hosting and processing scientific research data. The EOSC offers a web catalogue of computational infrastructures, analysis tools/platforms, and open data projects to match researchers with resource providers. It aims to establish scientific communities and research infrastructures towards (i) seamless access, (ii) data management from the perspective of FAIR (Findability, Accessibility, Interoperability, and Reusability)~\cite{wilkinson2016fair}, and (iii) reliable reuse of research data and all other digital objects produced along the research life cycle.

\smallskip
\noindent 
\textbf{Academic Clouds:} A project highly related to our work is EGI-ACE by European Grid Infrastructure (EGI)~\cite{egi}, which delivers cloud computing resources and services for European researchers. EGI-ACE aims to provide an open, data-centric, distributed, hybrid, and secure infrastructure, called the EOSC Compute Platform, which consists of computing and storage providers and platform services to support research and open science via data spaces. Similar to mdx, EGI-ACE is a data-science cloud service that includes low-level infrastructures (IaaS) as well as high-level services and software (PaaS, SaaS). 
Jetstream2~\cite{jetstream2} is an academic cloud service which includes a high-performance interactive VM environment (IaaS), secure data movement using Globus Transfer~\cite{chard2016globus} between multiple clouds, and data analysis resources. Jetstream2 consists of multiple campus-based cyberinfrastructures jointly connected through Internet2~\cite{internet2}.

The key difference between mdx and these projects is that they focus only on offering computing resources and associated software and services, whereas the mdx platform provides not only such resources, software, and services but also seamless and secure connection to data sources such as IoT devices and scientific facilities. To achieve this, the mdx network is designed to enhance the connectivity of SINET~\cite{sinet5}, which is an ultra high-speed/low-latency academic backbone network throughout Japan.

\smallskip
\noindent 
\textbf{Enterprise Clouds:} Enterprise clouds, such as Microsoft Azure~\cite{azure} and Google Cloud Platform (GCP)~\cite{gcp}, provide the integrated services of cloud computing and high-performance computing (e.g., a dedicated Cray XC/CS in Azure and custom HPC VMs in GCP) as well as IoT device management services (Azure IoT, Google Cloud IoT). 

Although most of the functions in mdx are technically feasible by a combination of these enterprise services, mdx is capable of more seamless and secure integration with world-class supercomputers thanks to its architecture, which was co-designed with SINET. Moreover, because mdx must manage cloud resources in a non-profit fashion, the simple resource management model of commercial clouds, with their pay-as-you-go services, does not fit the full set of technical challenges mdx must meet. 

\section{The \textit{mdx} platform}\label{mdx}

\begin{figure}
\centering
\includegraphics[width=.5\textwidth]{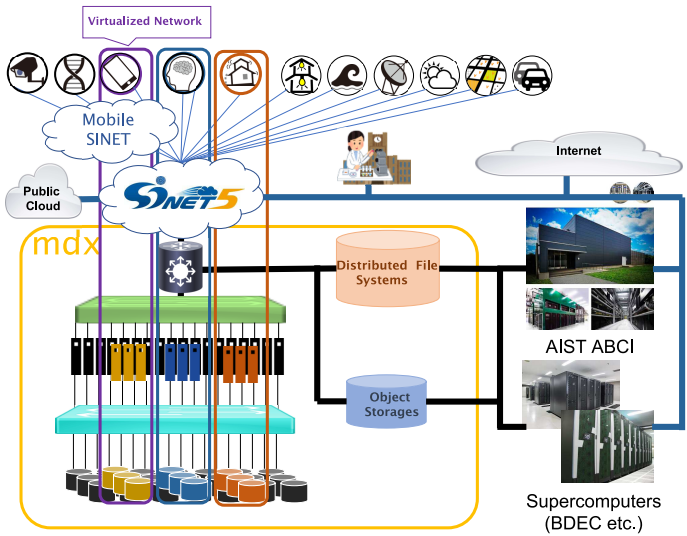}
\caption{The mdx Platform Overview.}
\label{fig:overview}
\end{figure}

In this section we describe how we can realize our vision described in Section \ref{motivation} and highlight important aspects of the platform---virtualization, data integration, isolation and security management, integration with supercomputers, authentication, open data platform, and community activities.
Figure~\ref{fig:overview} illustrates an overview of the platform.

\subsection{Virtualization and Hardware Overview}
\label{mdx:virt}

Virtualization is provided with the bare-metal hypervisor VMware ESXi in the VMware vSphere virtualization platform. VMware ESXi runs on top and accesses the hardware directly without the need to install an operating system. This direct access to hardware achieves minimizing overhead than other types of hypervisors. All of virtual machines are managed by VMware vSphere~\cite{vsphere} as the aggregated computing infrastructure, including CPU, storage, and network resources. 
As for the underlying hardware infrastructure, the platform is equipped with 368 CPU nodes and 40 GPU nodes as computing resources, and 1.0 Petabyte NVMe disk storage and 16.3 Petabyte hard disk for the Lustre file system, and 10.3 Petabyte S3-compliant object storage (DDN S3 Data Services) as storage resources. Each compute node employs 2 sockets of Intel Xeon Platinum 8368 with 38 cores of 2.4 GHz clock, and each GPU node contains 8 GPUs of NVIDIA A100. The network specification and features are mentioned in Section~\ref{mdx:network}.

\subsection{Integration with Data Sources}
\label{mdx:data}
In addition to providing a high-performance interactive computing environment, providing high-speed low-latency secure network connections to various data sources throughout Japan (e.g. IoT and mobile edge devices, laboratory instruments, and national scientific facilities) is an important feature of the mdx platform, enabling it to perform large-scale real-time analysis in a secure way. To enhance network performance and security, the mdx network is co-designed with the existing ultra high-speed/low-latency academic backbone network in Japan, referred to as SINET~\cite{sinet5}.

SINET is a national Research and Education Network or REN in Japan. It connects over 900 institutions, including all Japanese national universities, with other RENs such as Internet2 and G\`EANT, and with the Internet using 100 Gbps-based circuits, which will be migrated to 400 Gbps-based ones in April 2022. Japanese supercomputers, such as Fugaku~\cite{sato2020co}, are also connected to SINET. In addition to SINET's high-speed and broad coverage, a key feature of the network is its nationwide L2VPN (Layer2 Virtual Private Network) service. Institutions connected to SINET can deploy their private LANs (Local Area of Networks) over SINET with strong network isolation. An example of the SINET L2VPN service is Japan Data eXchange Network (JDXnet) for earthquake observation. Earthquake observation data flows through a SINET L2VPN deployed among about a dozen of universities and research institutions.  In 2018, SINET extended its VPN capability to mobile networks, called Mobile SINET~\cite{sinetstream}. IoT devices equipped with Mobile SINET SIM cards directly connect to SINET VPNs that can be extended to the institutions while keeping network isolation. Thus, SINET provides high-speed access to various research facilities---experimental equipment in the institutions, supercomputers, and IoT devices---while preserving security.

\subsection{Isolation and Security Management}
\label{mdx:network}
Data sources spread across the nation are accessible through SINET. The goal of the intra-network in the mdx platform is to extend this accessibility to VMs while achieving high-performance virtual networking for computation and strong isolation for security. To this end, we built two different types of networks: a flexible external network and a high-speed internal network.

The external network connects VMs with SINET. VMs can access the Internet through SINET with network address translation (NAT). The NAT function also exposes VMs to the Internet by mapping a global IPv4 address to a VM so that any devices on the Internet can access the VM. Users can configure packet filters to protect VMs exposed to the Internet. In addition to providing basic functions like cloud networking, the key point of the external network is that it is capable of extending layer-2 networks from SINET L2VPN to arbitrary VMs. VMs in the mdx platform can directly access the data sources through the virtualized layer-2 capability of the network and SINET L2VPN with strong security isolation. Such layer-2 networks can be closed---not accessible from other networks, including the Internet. The mdx platform is currently connected to SINET with two 100Gbps links, and we will migrate them into two 400Gbps links in 2022.

The internal network connects VMs without external accessibility. Instead, it provides high-bandwidth and low-latency compared to the external network. The internal network is a full-bisection spine-leaf topology where hypervisors are connected with 100Gbps links. The network supports Remote Direct Memory Access (RDMA) to enhance performance with the high-speed links. Users can run RDMA-capable applications such as MPI (Message Passing Interface) across VMs in the mdx platform. Storage systems are also connected to this network so that VMs access them over low-latency RDMA.

Because of scalability issues, the recent trend for large-scale data center networks is to build the network in layer-3 with Border Gateway Protocol~\cite{rfc7938}. However, the demands of the mdx network---extending SINET L2VPN to arbitrary VMs and user network isolation---require a network in layer-2 for VMs. To achieve flexible and scalable layer-2 networks, we adopted Virtual eXtensible LAN (VXLAN).  VXLAN is a protocol for Ethernet over IP overlay. The underlay of the external and internal networks are layer-3 networks with spine-leaf topology, and user networks are isolated by layer-2 segments (VLAN). Extending the user VLANs over the layer-3 underlays with VXLAN achieves flexibility and scalability simultaneously. In the external network, a segment from SINET L2VPN is carried to VMs over VXLAN. In the internal network, RDMA traffic is also carried over VXLAN thanks to RDMA over Converged Ethernet (RoCE). InfiniBand, a popular interconnect supporting RDMA, also has a feature for partitioning networks like pKeys, but the feature is not sufficient to prevent malicious users from breaking the isolation. Therefore, we adopted VLANs (Virtual LANs) to isolate users in the network and support RDMA with RoCE. Moreover, VXLAN enables carrying any RoCE segments to arbitrary VMs over the layer-3 full-bisection network.

\subsection{Integration with Supercomputers}
\label{mdx:supercomputers}
The mdx platform offers interactive high-performance VMs, but when workloads become much larger-scale and time-consuming, coupling with supercomputers may be an alternative solution to scale out the workloads.
The mdx platform has the high-performance/low-latency inter-networks of university-based supercomputers and top-tier supercomputers in Japan (e.g., Fugaku~\cite{sato2020co}). By effectively utilizing these networks, the mdx platform can tightly couple with these supercomputers. 
For simple  data sharing, the mdx provides S3-compliant object storage.
For further collaboration under seamless data management, mdx's internal storages can be directly applicable by L2VPN on SINET from  computing nodes in the supercomputers without proxy controls.
Moreover, for migrating applications on the mdx platform to supercomputers, the mdx platform needs to support efficient and easy-to-use HPC containerization such as  Singularity and Apptainer~\cite{kurtzer2017singularity,apptainer}.

\subsection{Authentication}
\label{mdx:authentication} 
User management and authentication are crucial for mdx, as its service covers a wide range of users from different universities and organizations. In order to achieve Single Sign-On (SSO) and seamless integration with other services such as supercomputers and university’s local services, mdx is based on GakuNin~\cite{gakunin}, a widely-used federated authentication system named mainly for Japanese academic institutes. This is similar to EduGain~\cite{edugain} in Europe and InCommon~\cite{incommon} and in the United States.

\subsection{Data Sharing}
\label{mdx:datasharing} 
In order to accelerate cross-disciplinary collaborations, allowing users to share data and meet collaboration partners on the platform is of great importance. To that end, we paid special attention to the design of this feature. Even though open data and open science has been recognized as important concepts, data sharing policy still vary in domains. Data providers, who owns confidential, commercial, or privacy-sensitive data, can only share the data with only approved users. In some research domains, they would like to open data only to those who have already collaboration agreement. With these constraints, we have designed a functionality where data providers can share their data to the public or certain groups of researchers with privileged access permission. 


\subsection{Scheduling and Pricing for Resource Provisioning}
\label{mdx:scheduling}
The mdx platform is a non-profit service, i.e. the platform basically charges users only for minimum operation costs such as energy cost and maintenance cost.
In order to collect such operational costs fairly from users while offering equal opportunities to a wide range of the users to access the resources (i.e., avoiding resource occupation by few users), mdx's scheduling and pricing policy for resource provision should be open and include two features as follows:

\smallskip
\noindent
\textbf{Configurable Spot/non-Spot Instance Management:}
\label{mdx:spot}
In mdx, all VM resources are managed by two types: spot and non-spot instance.
A spot instance is an on-demand virtualization instance which does not have lifetime guarantee - meaning that it could be killed within 24 hours if the platform resources are fully occupied. It is typically used for short-term purposes such as AI model training and web-cache servers.
A non-spot instance provides exclusive and reserved virtualization service. It is used for software that needs to continuously provide their services to users such as databases, real-time systems, and web servers.

We make the spot/non-spot instance ratio over the platform configurable in such a way that the administrators can manage occupied or idle VM resources and control each user's availability by referring to overall platform's situations and needs, e.g., system capacity, total energy usage, and so forth.

\smallskip
\noindent
\textbf{Offloading Workloads to Supercomputers:}
\label{mdx:offload}
The other functionality for managing occupied resources on the mdx platform is offloading workloads to supercomputers.
By providing a seamless connection to supercomputers (as discussed in~\ref{mdx:supercomputers}), the mdx platform avoids resource occupation caused by the large-scale batch-processing jobs, which are more suitable for supercomputers.

\subsection{Community Activities}
\label{mdx:community}
The mdx platform is an open platform managed by the 9 universities and 2 national research institutes in Japan.
Though the mdx platform, we aim to establish an open and cross-disciplinary research community.

\smallskip
\noindent
\textbf{Software:} 
The core components of the mdx platform are comprised of open source software such as operating systems, distributed file systems, networking tools, cluster managements, collaboration tools, and so on.
The information on the usage of these softwares on mdx is documented, and common software with typical configurations are shared among mdx users. 
For example, mdx offers VM templates which includes fundamental software components to fully utilize the functionalities provided by the mdx platform. 
Since the mdx platform provides virtual machine services, users need to install an OS on each virtual machine and install and configure GPU and network drivers and libraries such as Lustre to use the cluster hardware.
Such OS settings are non-essential tasks especially for application users, so the mdx platform provides an installed VM image in the mdx web portal.
Users select an installed OS image from the mdx web portal and deploy the research  environment.
The OS images in the current VM templates are a simple Linux OS with minimal security settings and drivers and libraries installed.
In the future, we plan to add the features for users to catalog and share their OS image between users.

\smallskip
\noindent
\textbf{Proposal-based Resource Provisioning:} 
Even though we plan to offer a paid service, we also offer mdx resources without any financial charge to cultivate more cross-disciplinary researches and support young/women scientists through peer-reviewed research proposals such as Joint Usage/Research Center for Interdisciplinary Large-scale Information Infrastructures (JHPCN)~\cite{jhpcn}. As the first trial in 2022, 15 research projects have been approved and will start to use the mdx platform from April 2022 for one year. Accepted research projects vary in a broad range of domains from earth science, historical science, materials science, biomass energy science, economics, job matching in nursing and medical domains, weather prediction, and so forth.

\smallskip
\noindent
\textbf{Seminars and Education:} 
We regularly hold research seminars related to data science for mdx users to accelerate the cross-disciplinary researches.
We will also hold educational seminars for students or non-professional users to learn how to use mdx.


\section{Project Status}\label{status}
As of this writing,  virtualization in Section \ref{mdx:virt}, network design in Section \ref{mdx:network}, authentication in Section \ref{mdx:authentication}, and integration with data in Section \ref{mdx:data} have been implemented and in operation since Fall 2021. We are currently in the midst of implementing the spot instance features in Section \ref{mdx:spot}, and the fair resource scheduling mechanism in Section \ref{mdx:scheduling} is planned to be available in Summer 2022. Some community-based approaches in Section \ref{mdx:community} have been started.  In the year 2023 and beyond, more functionalities, including data sharing in Section \ref{mdx:datasharing}, integration with supercomputers in Section \ref{mdx:supercomputers} and offloading workloads to supercomputers in Section \ref{mdx:offload}, will be designed and implemented. 

At this stage of mdx's development users can apply for a project by specifying its purpose, the number of CPUs and GPUs, disk size for virtual machines,  storage type (SSD, HDD, object storage), and global IPs. Once the project is approved, the specified resources are reserved so that users can promptly start deploying virtual machines either using VM templates that we prepared or using an ISO image of their desired operating system.  The project might be rejected if the claimed resources are beyond available resources or if the purpose of the project does not match with the usage policy of the mdx platform. Currently the mdx platform can not be used for commercial purposes.   
At this stage, more than 50 projects have been already approved. The areas of the projects are diverse such as materials discovery, genome analysis, satellite image analysis, biomass science, geo-spatial data analysis for smart cities. We anticipate more projects to come.  


\section{Concluding Remarks and Future Directions}
\label{summary}
In this paper, we described the vision of the mdx platform that aims at building a nation-wide cloud platform for data science that brings flexibility, agility, openness, and tight coupling with supercomputers and edge devices. The mdx platform has been fully in operation since Fall 2021 and 50+ projects have been using it. Through the mdx platform, we anticipate more collaborations for such problems that need interdisciplinary approaches by bringing knowledge and skills in data science and application domains.  As for our future plan, we will implement more functionalities that are listed up in the paper to achieve our vision, put them in production, and continuously evaluate and refine the platform by obtaining quantitative metrics in system usage as well as users’ feedbacks. We plan to continuously update the web site of the mdx project at~\cite{mdx-web}.

\balance
\bibliographystyle{IEEEtran}
\bibliography{ref}

\end{document}

%% file: authors.tex
\author{\IEEEauthorblockN{
        Toyotaro Suzumura\IEEEauthorrefmark{1},
        Akiyoshi Sugiki\IEEEauthorrefmark{2},
        Hiroyuki Takizawa\IEEEauthorrefmark{3},
        Akira Imakura\IEEEauthorrefmark{4},
        Hiroshi Nakamura\IEEEauthorrefmark{1},
    }
    \IEEEauthorblockN{
        Kenjiro Taura\IEEEauthorrefmark{1}, 
        Tomohiro Kudoh\IEEEauthorrefmark{1}, 
        Toshihiro Hanawa\IEEEauthorrefmark{1}, 
        Yuji Sekiya\IEEEauthorrefmark{1},
        Hiroki Kobayashi\IEEEauthorrefmark{1},
    }
    \IEEEauthorblockN{
        Shin Matsushima\IEEEauthorrefmark{1}, 
        Yohei Kuga\IEEEauthorrefmark{1},
        Ryo Nakamura\IEEEauthorrefmark{1}, 
        Renhe Jiang\IEEEauthorrefmark{1},
        Junya Kawase\IEEEauthorrefmark{1}, 
    }
    \IEEEauthorblockN{
        Masatoshi Hanai\IEEEauthorrefmark{1},        
        Hiroshi Miyazaki\IEEEauthorrefmark{5},
        Tsutomu Ishizaki\IEEEauthorrefmark{5},
        Daisuke Shimotoku\IEEEauthorrefmark{5},
        Daisuke Miyamoto\IEEEauthorrefmark{5},
    }
    \IEEEauthorblockN{
        Kento Aida\IEEEauthorrefmark{7},
        Atsuko Takefusa\IEEEauthorrefmark{7}, 
        Takashi Kurimoto\IEEEauthorrefmark{9}, 
        Koji Sasayama\IEEEauthorrefmark{9},
        Naoya Kitagawa\IEEEauthorrefmark{9}, 
    }
    \IEEEauthorblockN{
        Ikki Fujiwara\IEEEauthorrefmark{10},
        Yusuke Tanimura\IEEEauthorrefmark{11},
        Takayuki Aoki\IEEEauthorrefmark{12}, 
        Toshio Endo\IEEEauthorrefmark{12}, 
        Satoshi Ohshima\IEEEauthorrefmark{13},
    }
    \IEEEauthorblockN{
        Keiichiro Fukazawa\IEEEauthorrefmark{14}, 
        Susumu Date\IEEEauthorrefmark{15}, 
        Toshihiro Uchibayashi\IEEEauthorrefmark{16}
    }
   
  \IEEEauthorblockA{
    \IEEEauthorrefmark{1}\textit{Information Technology Center, The University of Tokyo} \\
    \IEEEauthorrefmark{2}\textit{Information Initiative Center, Hokkaido University} \\
    \IEEEauthorrefmark{3}\textit{Cyberscience Center, Tohoku University} \\
    \IEEEauthorrefmark{4}\textit{Faculty of Engineering, Information and Systems, University of Tsukuba} \\
    \IEEEauthorrefmark{5}\textit{Division for Information and Communication Systems, The University of Tokyo} \\
    \IEEEauthorrefmark{6}\textit{Center for Spatial Information Science, The University of Tokyo} \\
    \IEEEauthorrefmark{7}\textit{Information Systems Architecture Science Research Division, National Institute of Informatics} \\
    \IEEEauthorrefmark{8}\textit{Center for Cloud Research and Development, National Institute of Informatics} \\
    \IEEEauthorrefmark{9}\textit{Research and Development Center for Academic Networks, National Institute of Informatics} \\
    \IEEEauthorrefmark{10}\textit{Research Center for Open Science and Data Platform, National Institute of Informatics} \\
    \IEEEauthorrefmark{11}\textit{Digital Architecture Research Center, National Institute of Advanced Industrial Science and Technology} \\
    \IEEEauthorrefmark{12}\textit{Global Scientific Information and Computing Center, Tokyo Institute of Technology} \\
    \IEEEauthorrefmark{13}\textit{Information Technology Center, Nagoya University} \\
    \IEEEauthorrefmark{14}\textit{Academic Center for Computing and Media Studies, Kyoto University} \\
    \IEEEauthorrefmark{15}\textit{Cybermedia Center, Osaka University} \\
    \IEEEauthorrefmark{16}\textit{Research Institute for Information Technology, Kyushu University} \\
  }
  \IEEEauthorblockA{suzumura@ds.itc.u-tokyo.ac.jp}
}